# Super-FAN: Integrated facial landmark localization and super-resolution of real-world low resolution faces in arbitrary poses with GANs

Adrian Bulat and Georgios Tzimiropoulos
Computer Vision Laboratory, The University of Nottingham, United Kingdom
{adrian.bulat, yorgos.tzimiropoulos}@nottingham.ac.uk

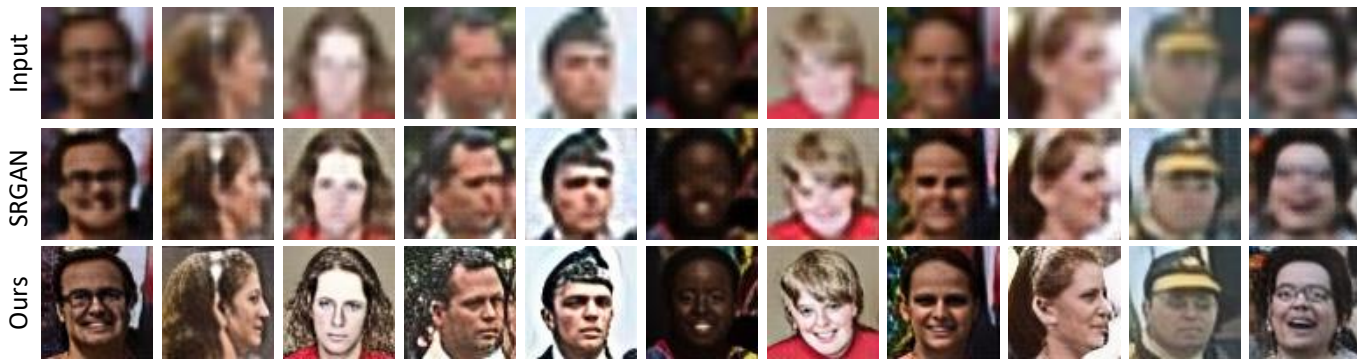

Figure 1: A few examples of visual results produced by our system on real-world low resolution faces from WiderFace.

## Abstract

*This paper addresses 2 challenging tasks: improving the quality of low resolution facial images and accurately locating the facial landmarks on such poor resolution images. To this end, we make the following 5 contributions: (a) we propose Super-FAN: the very first end-to-end system that addresses both tasks simultaneously, i.e. both improves face resolution and detects the facial landmarks. The novelty or Super-FAN lies in incorporating structural information in a GAN-based super-resolution algorithm via integrating a sub-network for face alignment through heatmap regression and optimizing a novel heatmap loss. (b) We illustrate the benefit of training the two networks jointly by reporting good results not only on frontal images (as in prior work) but on the whole spectrum of facial poses, and not only on synthetic low resolution images (as in prior work) but also on real-world images. (c) We improve upon the state-of-the-art in face super-resolution by proposing a new residual-based architecture. (d) Quantitatively, we show large improvement over the state-of-the-art for both face super-resolution and alignment. (e) Qualitatively, we show for the first time good results on real-world low resolution images like the ones of Fig. 1.*

## 1. Introduction

The aim of this paper is to improve upon the quality and understanding of very low resolution facial images. This is important in many applications, like face editing surveillance/security. In terms of quality, our aim is to increase the resolution and recover the details of real-world low resolution facial images like the ones shown in the first row of Fig. 1; this task is also known as face super-resolution (when the input resolution is too small this task is sometimes called face hallucination). In terms of understanding, we wish to extract mid- and high-level facial information by localizing a set a predefined facial landmarks with semantic meaning like the tip of the nose, the corners of the eyes etc.; this task is also known as face alignment.

Attempting to address both tasks simultaneously is really a chicken-and-egg problem: On one hand, being able to detect the facial landmarks has already been shown beneficial for face super-resolution [34, 30]; however how to accomplish this for low resolution faces in arbitrary poses is still an open problem [4]. On the other hand, if one could effectively super-resolve low quality and low resolution faces across the whole spectrum of facial poses, then facial landmarks can be localized with high accuracy.

Because it is difficult to detect landmarks in very low resolution faces (as noticed in [33, 34] and validated in this work), prior super-resolution methods based on this idea [34, 30] produce blurry images with artifacts when the facial landmarks are poorly localized. Our main contribution is to show that actually one can jointly perform facial landmark localization and super-resolution even for very low resolution faces in completely arbitrary poses (e.g. profile images, see also Figs. 1 and 5).

In summary, **our contributions** are:

1. We propose Super-FAN: the very first end-to-end system that addresses face super-resolution and alignment simultaneously, via integrating a sub-network for facial landmark localization through heatmap regression into a GAN-based super-resolution network, and incorporating a novel heatmap loss. See also Fig. 2.
2. We show the benefit of training the two networks jointly on both synthetically generated and real-world low-resolution faces of arbitrary facial poses.
3. We also propose an improved residual-based architecture for super-resolution.
4. Quantitatively, we report, for the first time, results across the whole spectrum of facial poses on the LS3D-W dataset [4], and show large improvement over the state-of-the-art on both super-resolution and face alignment.
5. Qualitatively, we show, for the first time, good visual results on real-world low resolution facial images taken from the WiderFace dataset [31] (see Figs. 1 and 5).

## 2. Closely related work

This section reviews related work in image and face super-resolution, and facial landmark localization.

**Image super-resolution.** Early attempts on super-resolution using CNNs [6, 16] used standard $L_p$ losses for training which result in blurry super-resolved images. To alleviate this, rather than using an MSE over pixels (between the super-resolved and the ground truth HR image), the authors of [15] proposed an MSE over feature maps, coined perceptual loss. Notably, we also use a perceptual loss in our method. More recently, in [20], the authors presented a GAN-based [7] approach which uses a discriminator to differentiate between the super-resolved and the original HR images and the perceptual loss. In [26], a patch-based texture loss is proposed to improve reconstruction quality.

Notice that all the aforementioned image super-resolution methods can be applied to all types of images and hence do not incorporate face-specific information, as proposed in our work. Also, in most cases, the aim is to produce high-fidelity images given an image which is already of good resolution (usually $128 \times 128$) while face super-resolution methods typically report results on very low resolution faces ($16 \times 16$ or $32 \times 32$).

From all the above mentioned methods, our work is more closely related to [15] and [20]. In particular, one of our contributions is to describe an improved GAN-based architecture for super-resolution, which we used as a strong baseline on top of which we built our integrated face super-resolution and alignment network.

**Face super-resolution.** The recent work of [32] uses a GAN-based approach (like the one of [20] without the perceptual loss) to super-resolve very low-resolution faces. The method was shown to work well for frontal and pre-aligned faces taken from the CelebA dataset [21]. In [33], the same authors proposed a two-step decoder-encoder-decoder architecture which incorporates a spatial transformer network to undo translation, scale and rotation misalignments. Their method was tested on pre-aligned, synthetically generated LR images from the *frontal* dataset of CelebA [21]. Notably, our network does not try to undo misalignment but simply learns how to super-resolve, respecting at the same time the structure of the human face by integrating a landmark localization sub-network.

The closest work to our method is [34] which performs face super-resolution and dense facial correspondence in an alternating manner. Their algorithm was tested on the frontal faces of PubFig [18] and Helen [19] while few results on real images (4 in total) were also shown with less success. The main difference with our work is that, in [34], the dense correspondence algorithm is not based on neural networks, but on cascaded regression, is pre-learned disjointly from the super-resolution network and remains fixed. As such, [34] suffers from the same problem of having to detect landmarks on blurry faces which is particularly evident for the first iterations of the algorithm. On the contrary, we propose learning both super-resolution and facial landmark localization *jointly* in an *end-to-end* fashion, and use *just one shot* to jointly super-resolve the image and localize the facial landmarks. See Fig. 2. As we show, this results in large performance improvement and generates images of high fidelity across the whole spectrum of facial poses.

It is worth noting that we go beyond the state-of-the-art and rigorously evaluate super-resolution and facial landmark localization across facial pose both quantitatively and qualitatively. As opposed to prior work which primarily uses frontal datasets [33, 5, 13, 34, 32, 30] (e.g. CelebA, Helen, LFW, BioID) to report results, the low resolution images in our experiments were generated using the newly created LS3D-W balanced dataset [4] which contains an even number of facial images per facial pose. We also report qualitatively results on more than 200 real-world low resolution facial images taken from the WiderFace dataset [31]. To our knowledge, this is the most comprehensive evaluation of face super-resolution algorithms on real images.

**Face alignment.** A recent evaluation of face alignment [4] has shown that when resolution drops down to 30 pixels, the performance drop of a state-of-the-art network trained on standard facial resolution ($192 \times 192$) for medium and large poses is more than 15% and 30%, respectively. This result is one of the main motivations behind our work. As our aim is not to propose a new architecture for face alignment, we employed the Face Alignment Network (FAN) of [4], built by combining the Hourglass network of [22] with the residual block of [3]. As shown in [4], FAN provides excellent performance across the whole spectrum of facial poses for good resolution images. As we show in this paper, a FAN specifically trained to localize the landmarks in low resolution images performs poorly. One of our contributions is to show that a FAN when integrated and jointly trained with a super-resolution network can localize facial landmarks in low resolution images with high accuracy.

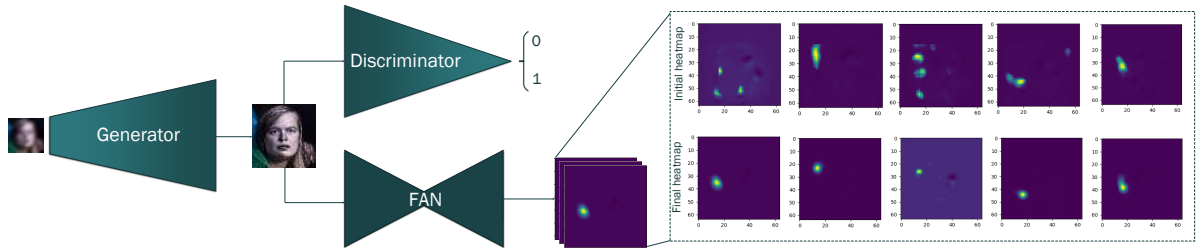

Figure 2: The proposed Super-FAN architecture comprises three connected networks: the first network is a newly proposed Super-resolution network (see sub-section 4.1). The second network is a WGAN-based discriminator used to distinguish between the super-resolved and the original HR image (see sub-section 4.2). The third network is FAN, a face alignment network for localizing the facial landmarks on the super-resolved facial image and improving super-resolution through a newly-introduced heatmap loss (see sub-section 4.3).

## 3. Datasets

To systematically evaluate face super-resolution across pose, we constructed a training dataset from 300W-LP [35], AFLW [17], Celeb-A [21] and a portion of LS3D-W balanced [4]. For testing, we used the remaining images from LS3D-W balanced, in which each pose range ($[0^o - 30^o], [30^o - 60^o], [60^o - 90^o]$) is equally represented.

**300W-LP** is a synthetically expanded dataset obtained by artificially rendering the faces from 300W [25] into large poses ($-90^0$ to $90^0$). While the dataset contains 61,225 images, there are only about 3,000 unique faces. Also, the images are affected by artifacts caused by the warping procedure. We included the entire dataset in our training set.

**AFLW** is a large-scale face alignment dataset that contains faces in various poses and expressions collected from Flickr. All 25,993 faces were included in our training set.

**Celeb-A** is a large-scale facial attribute dataset containing 10,177 unique identities and 202,599 facial images in total. Most of the images are occlusion-free and in frontal or near-frontal poses. To avoid biasing the training set towards frontal poses, we only used a randomly selected subset of approx. 20,000 faces.

**LS3D-W balanced** is a subset of the LS3D-W [4] dataset containing 7,200 images captured in-the-wild, in which each pose range ($[0^0 - 30^0], [30^0 - 60^0], [60^0 - 90^0]$) is equally represented (2,400 images each). We used 4,200 images for training, and kept 3000 for testing.

**WiderFace** is a face detection dataset containing 32,203 images with faces that exhibit a high degree of variability in pose, occlusion and quality. In order to assess the performance of our super-resolution method on in-the-wild, real-world images, we randomly selected 200 very low resolution, heavily blurred faces for qualitative evaluation.

## 4. Method

In this section, we describe the proposed architecture comprising of three connected networks: the first network is a Super-resolution network used to super-resolve the LR images. The second network is a discriminator used to distinguish between the super-resolved and the original HR images. The third network is FAN: the face alignment network for localizing the facial landmarks on the super-resolved facial images. Note that at test time the discriminator is not used. Overall, we call our network Super-FAN. See Fig. 2

Notably, for super-resolution, we propose a new architecture, shown in Fig. 3a, and detailed, along with the loss functions to train it, in sub-section 4.1. Our discriminator, based on Wasserstein GANs [1], is described in sub-section 4.2. Our integrated FAN along with our newly-introduced heatmap regression loss for super-resolution is described in sub-section 4.3. Sub-section 4.4 provides the overall loss for training Super-FAN. Finally, sub-section 4.5 describes the complete training procedure.

### 4.1. Super-resolution network

In this section, we propose a new residual-based architecture for super-resolution, inspired by [20], and provide the intuition and motivation behind our design choices. Our network as well as the one of [20] are shown in Figs. 3a and 3b, respectively. Their differences are detailed below. Following recent work [33, 32], the input and output resolutions are $16 \times 16$ and $64 \times 64$, respectively.

**Per-block layer distribution.** The architecture of [20], shown in Fig. 3b, uses 16, 1 and 1 blocks (layers) operating at the original, twice the original, and 4 times the original resolution, respectively; in particular, 16 blocks operate at a resolution 16 × 16, 1 at 32 × 32 and another 1 at 64 × 64. Let us denote this architecture as 16 − 1 − 1. We propose a generalized architecture of the form $N_1 - N_2 - N_3$, where $N_1, N_2$ and $N_3$ are the number of blocks used at the original, twice the original, and 4 times the original resolution, respectively. As opposed to the architecture of [20] where most of the blocks (i.e. 16) work at the input resolution, we opted for a more balanced distribution: 12-3-2, shown in Fig. 3a. Our motivation behind this change is as follows: since the main goal of the network is to super-resolve its input via hallucination, using only a single block at higher resolutions (as in [20]) is insufficient for the generation of sharp details, especially for images found in challenging scenarios (e.g. Fig. 1).

**Building block architecture.** While we experimented with a few variants of residual blocks [11, 12], similarly to [15, 20], we used the one proposed in [9]. The block contains two 3 × 3 convolutional layers, each of them followed by a batch normalization layer [14]. While [20] uses a PReLU activation function, in our experiments, we noticed no improvements compared to ReLU, therefore we used ReLUs throughout the network. See Fig. 3a.

**On the "long" skip connection.** The SR-ResNet of [20] groups its 16 modules operating at the original resolution in a large block, equipped with a skip connection that links the first and the last block, in an attempt to improve the gradient flow. We argue that the resolution increase is a gradual process in which each layer should improve upon the representation of the previous one, thus the infusion of lower level futures will have a small impact on the overall performance. In practice, and at least for our network, we found very small gains when using it. See supplementary material for additional results.

#### 4.1.1 Pixel and perceptual losses

**Pixel loss.** Given a low resolution image $I^{LR}$ (of resolution 16 × 16) and the corresponding high resolution image $I^{HR}$ (of resolution 64 × 64), we used the pixel-wise MSE loss to minimize the distance between the high resolution and the super-resolved image. It is defined as follows:

$$l_{pixel} = \frac{1}{r^2 WH} \sum_{x=1}^{rW} \sum_{y=1}^{rH} (I^{HR}_{x,y} - G_{\theta_G}(I^{LR})_{x,y})^2, \quad (1)$$

where $W$ and $H$ denote the size of $I^{LR}$ and $r$ is the upsampling factor (set to 4 in our case).

**Perceptual loss.** While the pixel-wise MSE loss achieves high PSNR values, it often results in images which lack fine details, are blurry and unrealistic (see Fig. 4). To address this, in [15, 20], a perceptual loss is proposed in which the super-resolved image and the original image must also be close in feature space. While [20] defines this loss over the activations of layer 5_4 (the one just before the FC layers) of VGG-19 [27], we instead used a combination of low, middle and high level features computed after the B1, B2 and B3 blocks of ResNet-50 [11]. The loss over the ResNet features at a given level $i$ is defined as:

$$l_{feature/i} = \frac{1}{W_i H_i} \sum_{x=1}^{W_i} \sum_{y=1}^{H_i} (\phi_i(I^{HR})_{x,y} - \phi_i(G_{\theta_G}(I^{LR}))_{x,y})^2, \quad (2)$$

where $\phi_i$ denotes the feature map obtained after the last convolutional layer of the $i-$th block and $W_i, H_i$ its size.

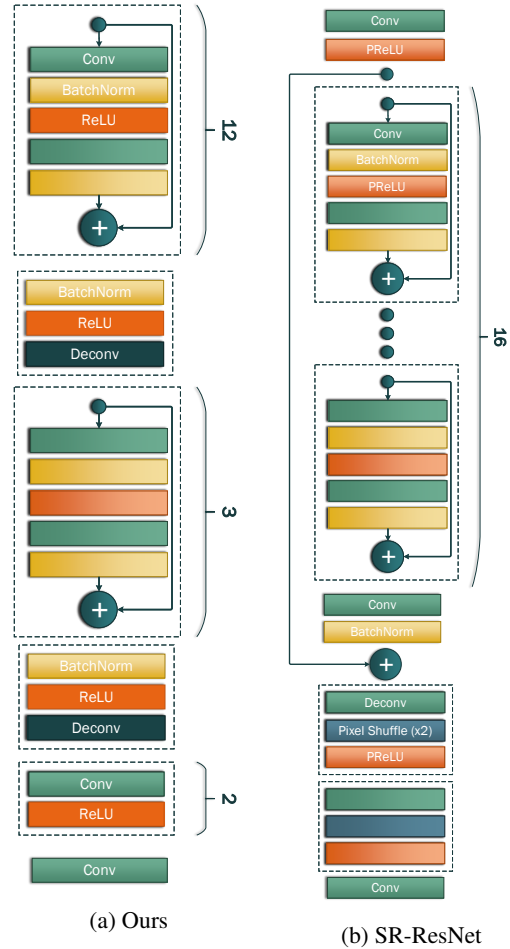

Figure 3: A comparison between the proposed super-resolution architecture (left) and the one described in [20] (right). See also sub-section 4.1.

### 4.2. Adversarial network

The idea of using a GAN [7] for face super-resolution is straightforward: the generator $G$ in this case is the super-resolution network which via a discriminator $D$ and an adversarial loss is enforced to produce more realistic super-resolved images lying in the manifold of facial images. Prior work in image super-resolution [20] used the GAN formulation of [24]. While in our work, we do not make an attempt to improve the GAN formulation per se, we are the first to make use of recent advances within super-resolution and replace [24] with the Wasserstein GAN of (WGAN) [1], as also improved in [10] (see also Eq. (3)).

We emphasize that our finding is that the improvement over [24] is only with respect to the stability and easiness of training and not with the quality of the super-resolved facial images: while training from scratch with the GAN loss of [24] is tricky and often leads to an unsatisfactory solution, by using a WGAN loss, we stabilized the training and allowed for the introduction of the GAN loss at earlier stages in the training process, thus reducing the overall training time. Finally, in terms of network architecture, we used the DCGAN [24] without batch normalization.

#### 4.2.1 Adversarial loss

Following [1] and [10], the WGAN loss employed in our face super-resolution network is defined as:

$$l_{WGAN} = \mathop{\mathbb{E}}_{\hat{I} \sim \mathbb{P}_g}[D(\hat{I})] - \mathop{\mathbb{E}}_{I \sim \mathbb{P}_r}[D(I^{HR})] \\ + \lambda \mathop{\mathbb{E}}_{\hat{I} \sim \mathbb{P}_{\hat{I}}}[\,(\|\nabla_{\hat{I}} D(\hat{I})\|_2 - 1)^2\,], \quad (3)$$

where $\mathbb{P}_r$ is the data distribution and $\mathbb{P}_g$ is the generator $G$ distribution defined by $\hat{I} = G(I^{LR})$. $\mathbb{P}_{\hat{I}}$ is obtained by uniformly sampling along straight lines between pairs of samples from $\mathbb{P}_r$ and $\mathbb{P}_g$.

### 4.3. Face Alignment Network

The losses defined above (pixel, perceptual and adversarial) have been used in general purpose super-resolution and although alone do provide descent results for facial super-resolution, they also fail to incorporate information related to the structure of the human face into the super-resolution process. We have observed that when these losses are used alone pose or expression related details may be missing or facial parts maybe incorrectly located (see Fig. 4).

To alleviate this, we propose to enforce facial structural consistency between the low and the high resolution image via integrating a network for facial landmark localization through heatmap regression into the super-resolution process and optimizing an appropriate heatmap loss.

To this end, we propose to use the super-resolved image as input to a FAN and train it so that it produces the same output as that of another FAN applied on the original high resolution image. We note that FAN uses the concept of heatmap regression to localize the landmarks: rather than training a network to regress a $68 \times 2$ vector of x and y coordinates, each landmark is represented by an output channel containing a 2D Gaussian centered at the landmark's location, and then the network is trained to regress the 2D Gaussians, also known as heatmaps. As a number of works have shown (e.g. [2]), these heatmaps capture shape information (e.g. pose and expression), spatial context and structural part relationships. Enforcing the super-resolved and the corresponding HR image to yield the same heatmaps via minimization of their distance is a key element of our approach: not only are we able to localize the facial landmarks but actually we impose these two images to have similar facial structure. In terms of architecture, we simply used FAN [4] with 2 Hourglass modules.

#### 4.3.1 Heatmap loss

Based on the above discussion, we propose to enforce structural consistency between the super-resolved and the corresponding HR facial image via a heatmap loss defined as:

$$l_{heatmap} = \frac{1}{N} \sum_{n=1}^{N} \sum_{ij} (\widetilde{M}_{i,j}^n - \widehat{M}_{i,j}^n)^2, \quad (4)$$

where $\widetilde{M}_{i,j}^n$ is the heatmap corresponding to the $n$-th landmark at pixel $(i, j)$ produced by running the FAN integrated into our super-resolution network on the super-resolved image $\hat{I}_{HR}$ and $\widehat{M}_{i,j}^n$ is the heatmap obtained by running another FAN on the original image $I_{HR}$.

Another key feature of our heatmap loss is that its optimization does not require having access to ground truth landmark annotations just access to a pre-trained FAN. This allows us to train the entire super-resolution network in a weakly supervised manner which is necessary since for some of the datasets used for training (e.g. CelebA) ground truth landmark annotations are not available, anyway.

### 4.4. Overall training loss

The overall loss used for training Super-FAN is:

$$l^{SR} = \alpha l_{pixel} + \beta l_{feature} + \gamma l_{heatmap} + \zeta l_{WGAN}, \quad (5)$$

where $\alpha, \beta, \gamma$ and $\zeta$ are the corresponding weights.

### 4.5. Training

All images were cropped based on the bounding box such that the face height is 50 px. Input and output resolutions were $16 \times 16$ px and $64 \times 64$ px, respectively. To avoid overfitting, we performed random image flipping, scaling (between 0.85 and 1.15), rotation (between $-30^o$ and $30^o$), color, brightness and contrast jittering. All models, except for the one trained with the GAN loss, were trained for 60 epochs, during which the learning rate was gradually decreased from 2.5e-4 to 1e-5. The model trained with the GAN loss was based on a previously trained model which was fine-tunned for 5 more epochs. The ratio between running the generator and the discriminator was kept to 1. Finally, for end-to-end training of the final model (i.e. Super-FAN), all networks (super-resolution, discriminator and FAN) were trained jointly for 5 epochs with a learning rate of 2.5e-4. All models, implemented in PyTorch [23], were trained using rmsprop [28].

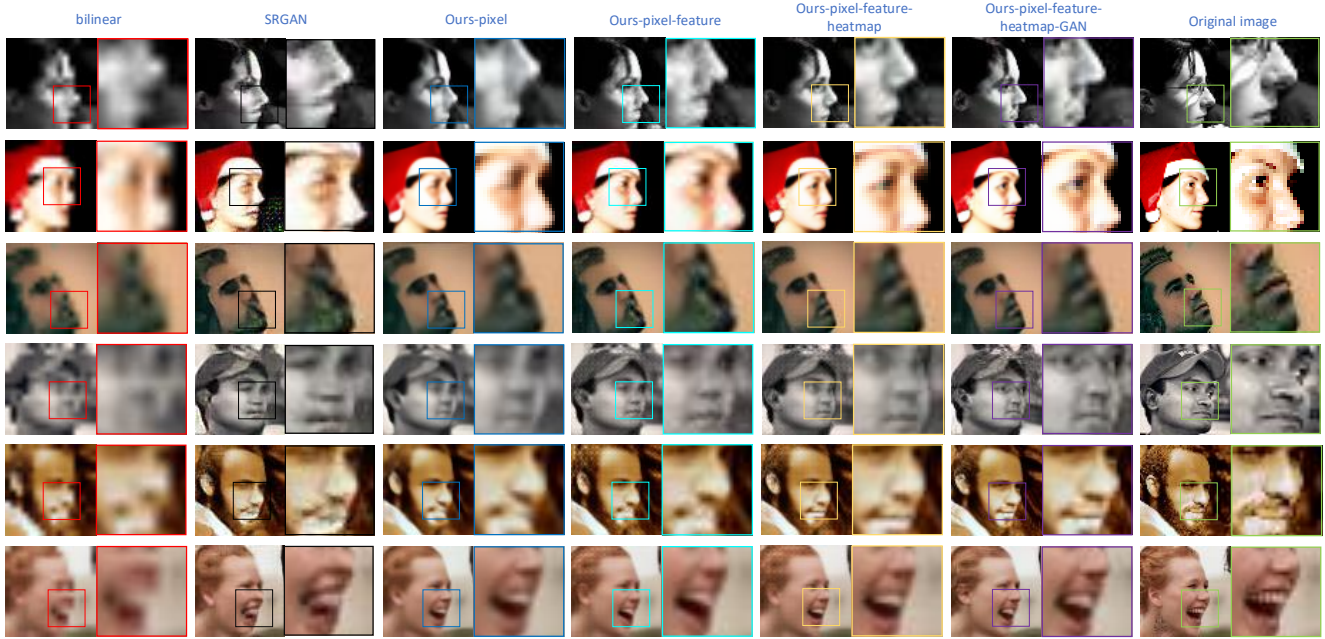

Figure 4: Visual results on LS3D-W. Notice that: (a) The proposed Ours-pixel-feature already provides better results than those of SR-GAN [20]. (b) By additionally adding the newly proposed heatmap loss (Ours-pixel-feature-heatmap) the generated faces are better structured and look far more realistic. Ours-pixel-feature-heatmap-GAN is Super-FAN which improves upon Ours-pixel-feature-heatmap by adding the GAN loss and by end-to-end training. Best viewed in electronic format.

## 5. Experiments

In this section, we evaluate the performance of Super-FAN. The details of our experiments are as follows:

**Training/Testing.** Unless otherwise stated, all methods, including [20], were trained on the training sets of section 3. We report quantitative and qualitative results on the subset of LS3D-W balanced consisting of 3,000 images, with each pose range being *equally* represented. We report qualitative results for more than 200 images from WiderFace.

**Performance metrics.** In sub-section 5.1, we report results using the standard super-resolution metrics, namely the PSNR and SSIM [29], confirming [20] that both of them are a poor measure of the perceived image quality. In sub-section 5.2, we report results on facial landmark localization accuracy. To alleviate the issues with PSNR and SSIM, we also propose another indirect way to assess super-resolution quality based on facial landmarks: in particular, we trained a FAN on high resolution images and then used it to localize the landmarks on the super-resolved images produced by each method. As our test set (LS3D-W balanced) provides the ground truth landmarks, we can use landmark localization accuracy to assess the quality of the super-resolved images: the rationale is that, the better the quality of the super-resolved image, the higher the localization accuracy will be, as the FAN used saw only real high resolution images during training. The metric used to quantify performance is the Area Under the Curve (AUC) [4].

**Variants compared.** In section 4, we presented a number of networks and losses for super-resolution which are all evaluated herein. These methods are named as follows:

- Ours-pixel: this is the super-resolution network of sub-section 4.1 trained with the pixel loss of Eq. (1).
- Ours-pixel-feature: this is the super-resolution network of sub-section 4.1 trained with the pixel loss of Eq. (1) and the perceptual loss of Eq. (2).
- Ours-pixel-feature-heatmap: this is the super-resolution network of sub-section 4.1 trained with the pixel loss of Eq. (1), the perceptual loss of Eq. (2), and the newly proposed heatmap loss of Eq. (4).
- Ours-Super-FAN: this improves upon ours-pixel-feature-heatmap by additionally training with the GAN loss of Eq. (3) and by end-to-end training.

**Comparison with the state-of-the-art.** We report results for the method of [20], implemented with and without the GAN loss, called SR-GAN and SR-ResNet, respectively, and for the standard baseline based on bilinear in-terpolation. We also show visual results on WiderFace by running the code from [34] [1].

## 5.1. Super-resolution results

Our quantitative results on LS3D-W across all facial poses are shown in Table 1. In terms of PSNR, the best results are achieved by Ours-pixel-feature-heatmap. In terms of SSIM, the best performing method seems to be Ours-pixel. From these numbers, it is hard to safely conclude which method is the best. Visually inspecting the super-resolved images though in Fig. 4 clearly shows that the sharper and more detailed facial images are by far produced by Ours-pixel-feature-heatmap and Ours-Super-FAN. Notably, Ours-pixel achieves top performance in terms of SSIM, yet the images generated by it are blurry and unrealistic (see Fig. 4), and are arguably less visually appealing than the ones produced by incorporating the other loss terms. We confirm the findings of [20] that these metrics can sometimes be misleading.

## 5.2. Facial landmark localization results

Herein, we present facial landmark localization results (on LS3D-W), also in light of our proposed way to evaluate super-resolution based on the accuracy of a pre-trained FAN on the super-resolved images (see **Performance metrics**). We report results for the following methods:

- FAN-bilinear: this method upsamples the LR image using bilinear interpolation and then runs FAN on it.
- Retrained FAN-bilinear: this is the same as FAN-bilinear. However, FAN was re-trained to work exclusively with bilinearly upsampled LR images.
- FAN-SR-ResNet: the LR image is super-resolved using SR-ResNet [20] and then FAN is run on it.
- FAN-SR-GAN: the LR image is super-resolved using SR-GAN [20] and then FAN is run on it.
- FAN-Ours-pixel: the LR image is super-resolved using Ours-pixel and then FAN is run on it.
- FAN-Ours-pixel-feature: the LR image is super-resolved using Ours-pixel-feature and then FAN is run on it.
- FAN-Ours-pixel-feature-heatmap-GAN: the LR image is super-resolved using Ours-pixel-feature-heatmap-GAN and then FAN is run on it. The FAN is **not trained** with the rest of the super-resolution network i.e. the same FAN as above was used. This variant is included to highlight the importance of jointly training the face alignment and super-resolution networks as proposed in this work.
- Super-FAN: this is the same as above however, this time, FAN is **jointly trained** with the rest of the network.
- FAN-HR images: this method uses directly the original HR images as input to FAN. This method provides an upper bound in performance.

The results are summarized in Fig. 4 and Table 2. See supplementary material for examples showing the landmark localization accuracy. From the results, we conclude that:

1. Super-FAN is by far the best performing method being the only method attaining performance close to the upper performance bound provided by FAN-HR images.
2. Jointly training the face alignment and super-resolution networks is necessary to obtain high performance: Super-FAN largely outperforms FAN-Ours-pixel-feature-heatmap-GAN (second best method).
3. The performance drop of Super-FAN for large poses ($> 60^o$) is almost twice as much as that of FAN-HR images. This indicates that facial pose is still an issue in face super-resolution.
4. Even a FAN trained exclusively to work with bilinearly upsampled images (Retrained FAN-Bilinear), clearly an unrealistic scenario, produces moderate results, and far inferior to the ones produced by Super-FAN.
5. FAN-Ours-pixel-feature outperforms both FAN-SR-GAN and FAN-SR-ResNet. This shows that the proposed super-resolution network of section 4.1 (which does not use heatmap or WGAN losses) already outperforms the state-of-the-art.
6. From FAN-Ours-pixel to Super-FAN, each of the losses added improves performance which is in accordance to the produced visual results of Fig. 4. This validates our approach to evaluate super-resolution performance indirectly using facial landmark localization accuracy.

## 5.3. Comparison on real-world images

Most face super-resolution methods show results on synthetically generated LR images. While these results are valuable for assessing performance, a critical aspect of any system is its performance on real-world data captured in unconstrained conditions. To address this, in this section and in our supplementary material, we provide visual results by running our system on more than 200 low resolution blurry images taken from the WiderFace and compare its performance with that of SR-GAN [20] and CBN [34].

Initially, we found that the performance of our method on real images, when trained on artificially downsampled images, was sub-optimal, with the super-resolved images often lacking sharp details. However, retraining Super-FAN by applying additionally random Gaussian blur (of kernel size between 3 and 7 px) to the input images, and simulating jpeg artifacts and color distortion, seems to largely alleviate the problem. Results of our method, SR-GAN (also retrained in the same way as our method) and CBN can be seen in Figs. 1 and 5, while the results on all 200 images can be found in the supplementary material.

---

[1]It is hard in general to compare with [34] because the provided code pre-processes the facial images very differently to our method.

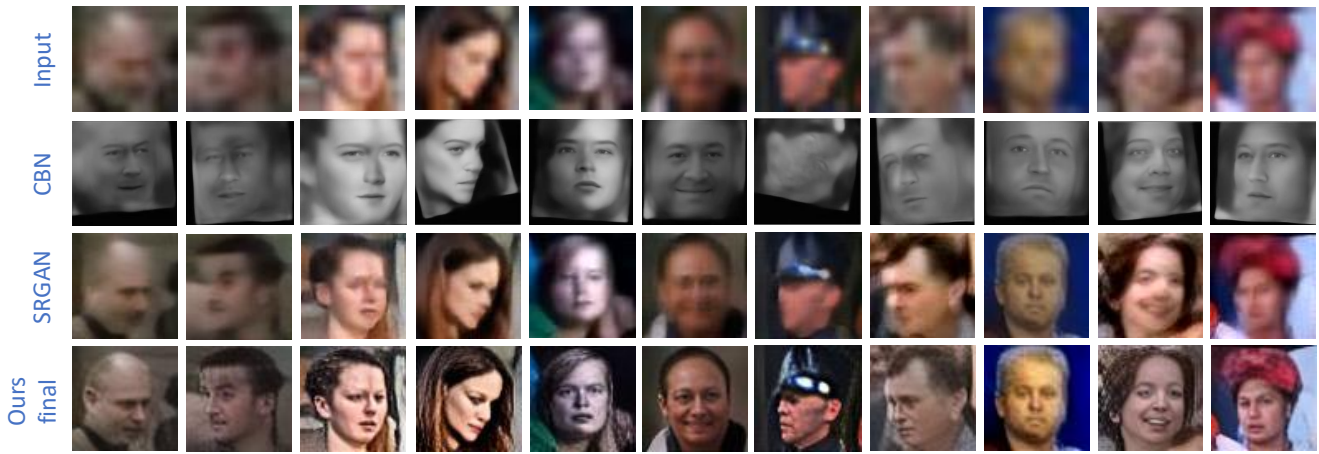

Figure 5: Results produced by our system, SR-GAN [20] and CBN [34] on real-world low resolution faces from WiderFace.

| Method | PSNR | | | SSIM | | |
|---|---|---|---|---|---|---|
| | 30 | 60 | 90 | 30 | 60 | 90 |
| bilinear upsample (baseline) | 20.25 | 21.45 | 22.10 | 0.7248 | 0.7618 | 0.7829 |
| SR-ResNet | 21.21 | 22.23 | 22.83 | 0.7764 | 0.7962 | 0.8077 |
| SR-GAN | 20.01 | 20.94 | 21.48 | 0.7269 | 0.7465 | 0.7586 |
| Ours-pixel | **21.55** | 22.45 | 23.05 | **0.8001** | **0.8127** | **0.8240** |
| Ours-pixel-feature | 21.50 | 22.51 | 23.10 | 0.7950 | 0.7970 | 0.8205 |
| Ours-pixel-feature-heatmap | **21.55** | **22.55** | **23.17** | 0.7960 | 0.8105 | 0.8210 |
| Ours-Super-FAN | 20.85 | 21.67 | 22.24 | 0.7745 | 0.7921 | 0.8025 |

Table 1: PSNR- and SSIM-based super-resolution performance on LS3D-W balanced dataset across pose (higher is better). The results are not indicative of visual quality. See Fig. 4.

| Method | [0-30] | [30-60] | [60-90] |
|---|---|---|---|
| FAN-bilinear | 10.7% | 6.9% | 2.3% |
| FAN-SR-ResNet | 48.9% | 38.9% | 21.4% |
| FAN-SR-GAN | 47.1% | 36.5% | 19.6% |
| Retrained FAN-bilinear | 55.9% | 49.2% | 37.8% |
| FAN-Ours-pixel | 52.3% | 45.3% | 28.3% |
| FAN-Ours-pixel-feature | 57.0% | 50.2% | 34.9% |
| FAN-Ours-pixel-feature-heatmap | 61.0% | 55.6% | 42.3% |
| **Super-FAN** | **67.0%** | **63.0%** | **52.5%** |
| FAN-HR images | 75.3% | 72.7% | 68.2% |

Table 2: AUC across pose (calculated for a threshold of 10%; see [4]) on our LS3D-W balanced test set. The results, in this case, are indicative of visual quality. See Fig. 4.

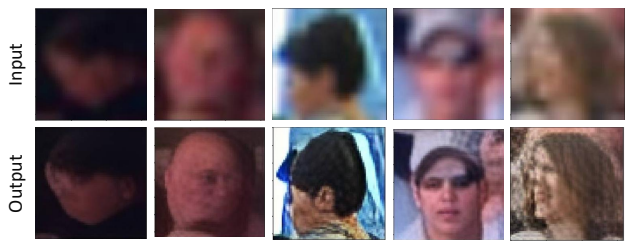

Figure 6: Failure cases of our method on WiderFace. Typically, these include extreme facial poses, large occlusions and heavy blurring.

Our method provides the sharper and more detailed results performing well across all poses. SR-GAN fails to produce sharp results. CBN produces unrealistic results especially for the images that landmark localization was poor.

A few failure cases of our method are shown in Fig. 6; mainly cases of extreme poses, large occlusions and heavy blurring. With respect to the latter, although our augmentation strategy seems effective, it is certainly far from optimal. Enhancing it is left for interesting future work.

## 6. Conclusions

We proposed Super-FAN: the very first end-to-end system for integrated facial super-resolution and landmark localization. Our method incorporates facial structural information in a newly proposed architecture for super-resolution, via integrating a sub-network for face alignment and optimizing a novel heatmap loss. We show large improvement over the state-of-the-art for both face super-resolution and alignment across the whole spectrum of facial poses. We also show, for the first time, good results on real-world low resolution facial images.

## A1. Ablation studies

This section describes a series of experiments, further analyzing the importance of particular components on the overall performance. It also provides additional qualitative results.

### A1.1. On the pixel loss

In this section, we compare the effect of replacing the L2 loss of Eq. 1 with the L1 loss. While the L1 loss is known to be more robust in the presence of outliers, we found no improvement of using it over the L2 loss. The results are shown in Table 3.

### A1.2. On the heatmap loss

Similarly to the above experiment, we also replaced the L2 heatmap loss of Eq. 4 with the L1 loss. The results are shown in Table 5, showing descent improvement for large poses.

### A1.3. On the importance of the skip connection

Herein, we analyzed the impact of the long-skip connections to the overall performance of the generator. The results, shown in Table 4, show no improvement.

### A1.4. On network speed

Besides accuracy, another important aspect of network performance is speed. Compared with SR-GAN [20], our generator is only 10% slower, being able to process 1,000 images in 4.6s (vs. 4.3s required by SR-GAN) on an NVIDIA Titan-X GPU.

### A1.5. Additional qualitative results

Fig. 9 shows the results produced by Super-FAN on all of the 200 randomly selected low-resolution images from WiderFace. Fig. 8 shows the face size distribution Notice that our method copes well with pose variation and challenging illumination conditions. There were a few failure cases, but in most of these cases, it is impossible to tell whether the low-resolution image was actually a face.

Fig. 10 shows a few fitting results produced by Super-FAN on the LS3D-W Balanced dataset. The predictions were plotted on top of the low resolution input images. We observe that our method is capable of producing accurate results even for faces found in arbitrary poses exhibiting various facial expressions.

We also tested our system on images from the Surveillance Cameras Face dataset (SCface) [8]. The dataset contains 4,160 images of 130 unique subjects taken with different cameras from different distances. Fig. 7 shows a few qualitative results from this dataset.

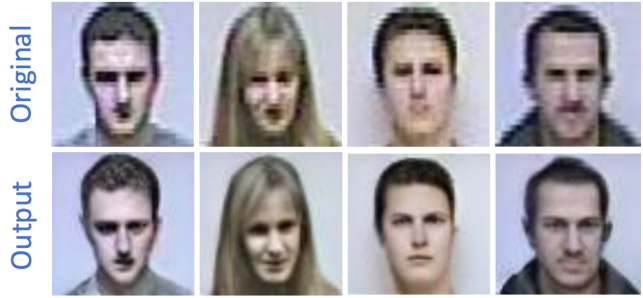

Figure 7: Qualitative results on the SCface dataset [8].

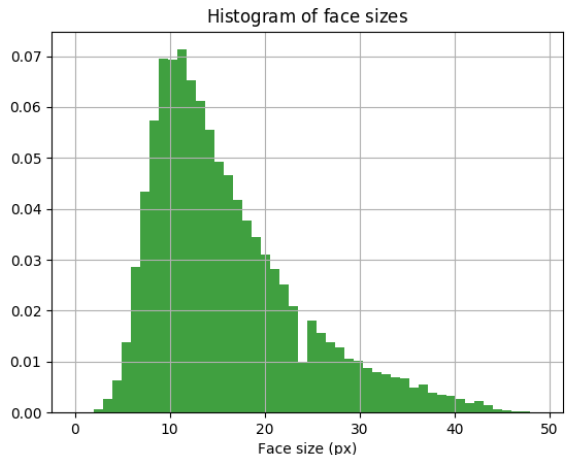

Figure 8: Face size (defined as $\max(width, height)$) distribution of the selected subset of low resolution images from WiderFace.

| Method | PSNR | | | SSIM | | |
|---|---|---|---|---|---|---|
| | 30 | 60 | 90 | 30 | 60 | 90 |
| Ours-pixel (L2) | 21.55 | 22.45 | 23.05 | 0.8001 | 0.8127 | 0.8240 |
| Ours-pixel (L1) | 21.47 | 22.40 | 23.00 | 0.7988 | 0.8120 | 0.8229 |

Table 3: PSNR and SSIM when training our generator with L2 and L1 pixel-losses.

| Method | PSNR | | | SSIM | | |
|---|---|---|---|---|---|---|
| | 30 | 60 | 90 | 30 | 60 | 90 |
| Ours-pixel (no-skip) | 21.55 | 22.45 | 23.05 | 0.8001 | 0.8127 | 0.8240 |
| Ours-pixel (with skip) | 21.56 | 22.45 | 23.04 | 0.8021 | 0.8132 | 0.8241 |

Table 4: PSNR and SSIM for "no-skip" and "with skip" versions. The "no-skip" version indicates the absence of the long skip connection (the network depicted in Fig. 3a), while the "with skip" version adds two new long skip connections, similarly to [11].

| Method | [0-30] | [30-60] | [60-90] |
|---|---|---|---|
| FAN-Ours-pixel-feature-heatmap (L2) | 61.0% | 55.6% | 42.3% |
| FAN-Ours-pixel-feature-heatmap (L1) | 61.1% | 55.4% | 42.0% |

Table 5: AUC across pose (on our LS3D-W balanced test set) for L2 and L1 heatmap losses.

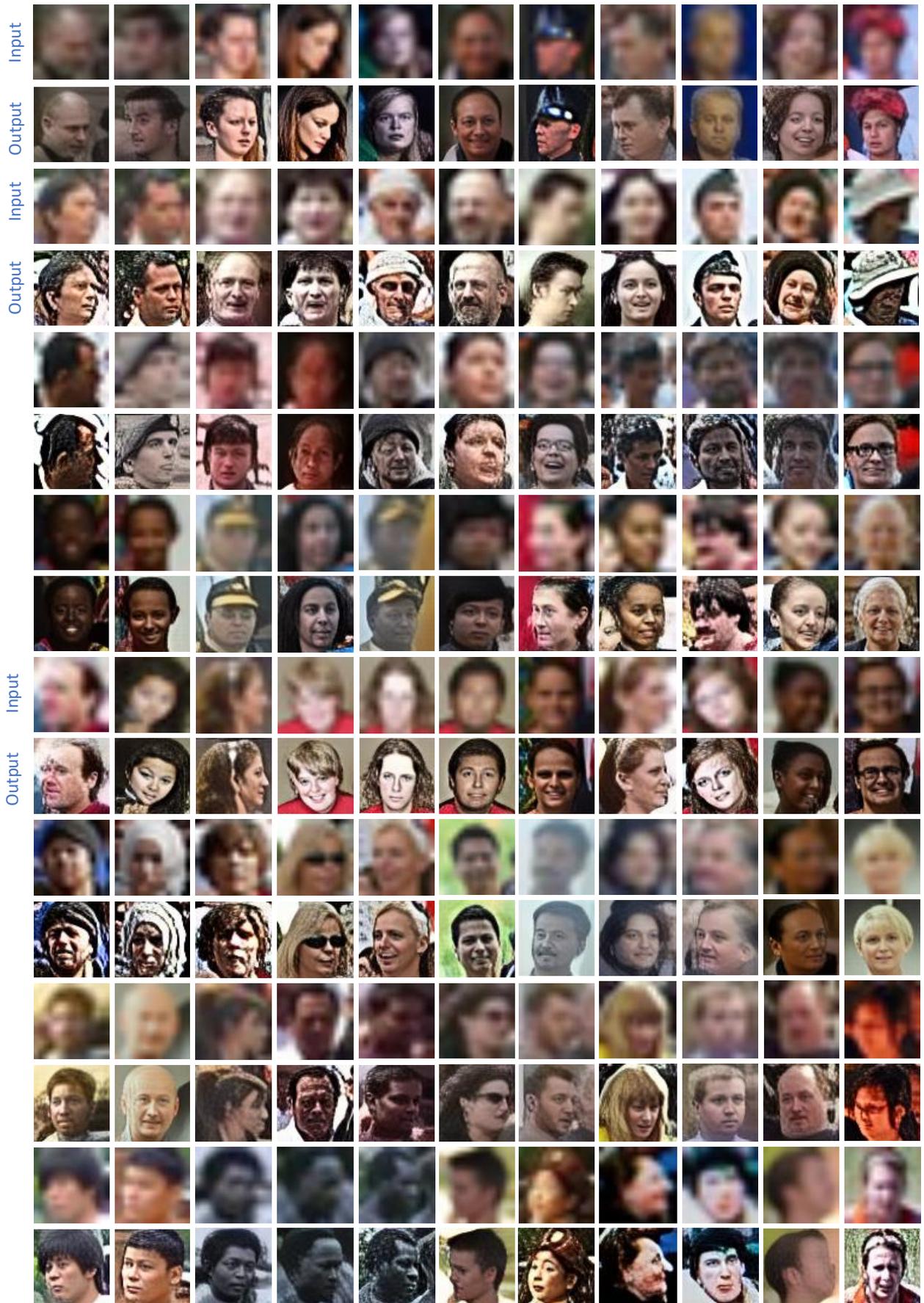

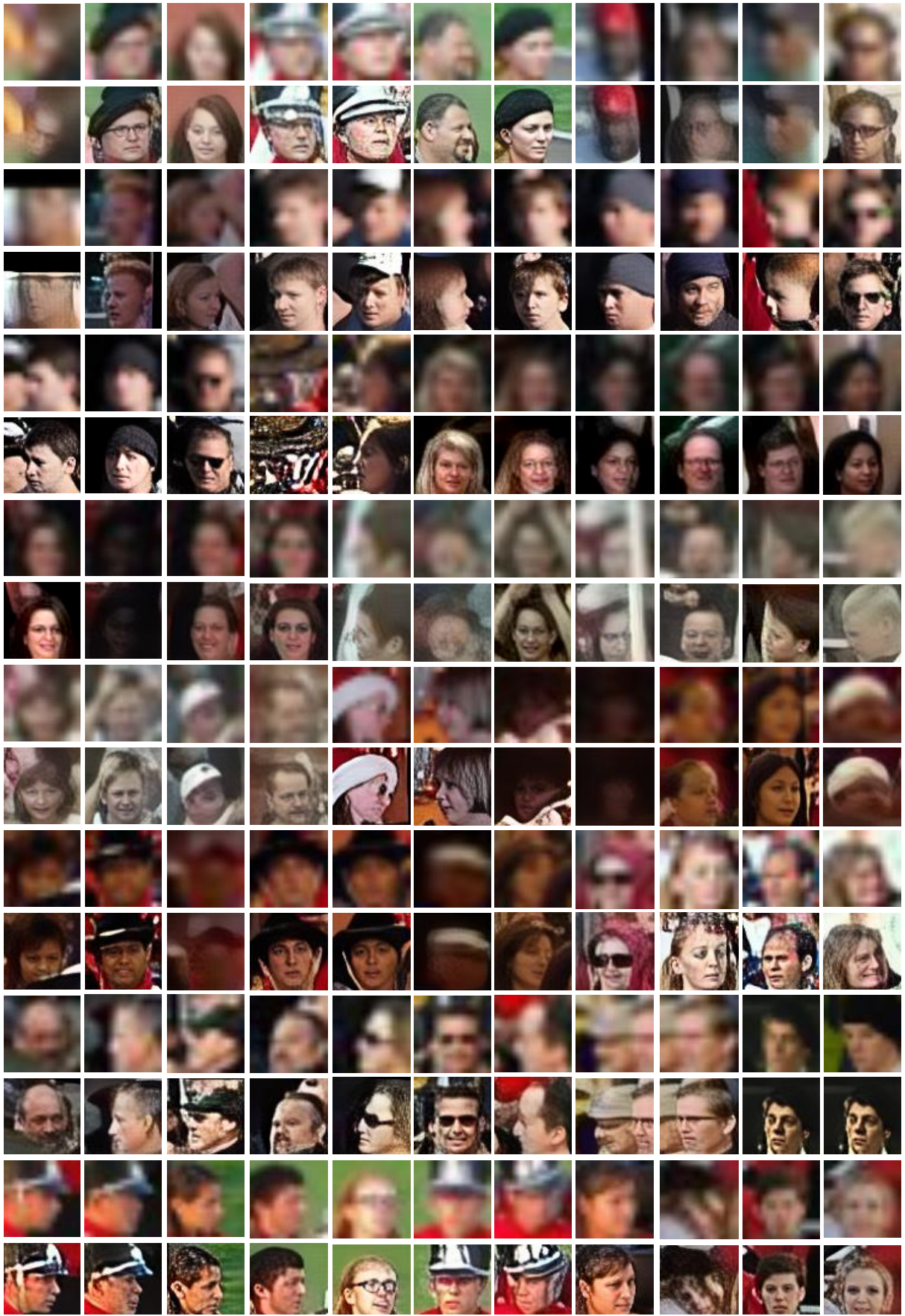

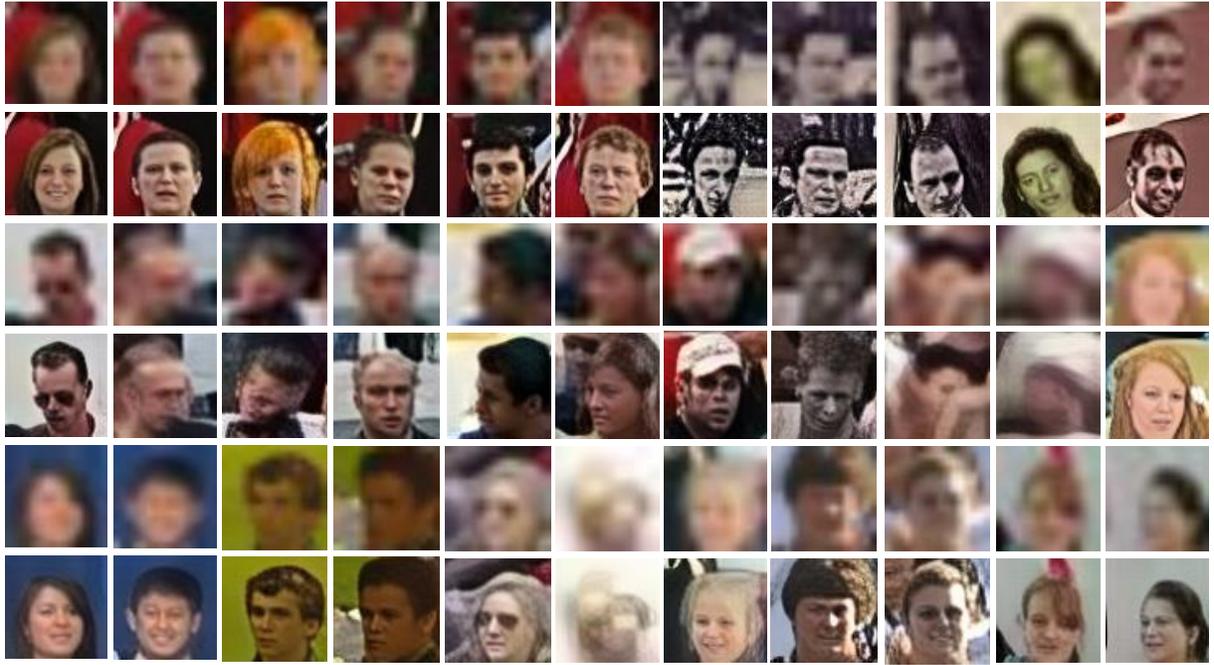

Figure 9: Visual results on a subset of very low resolution images from the WiderFace dataset. The odd rows represent the input, while the even ones the output produced by Super-FAN.

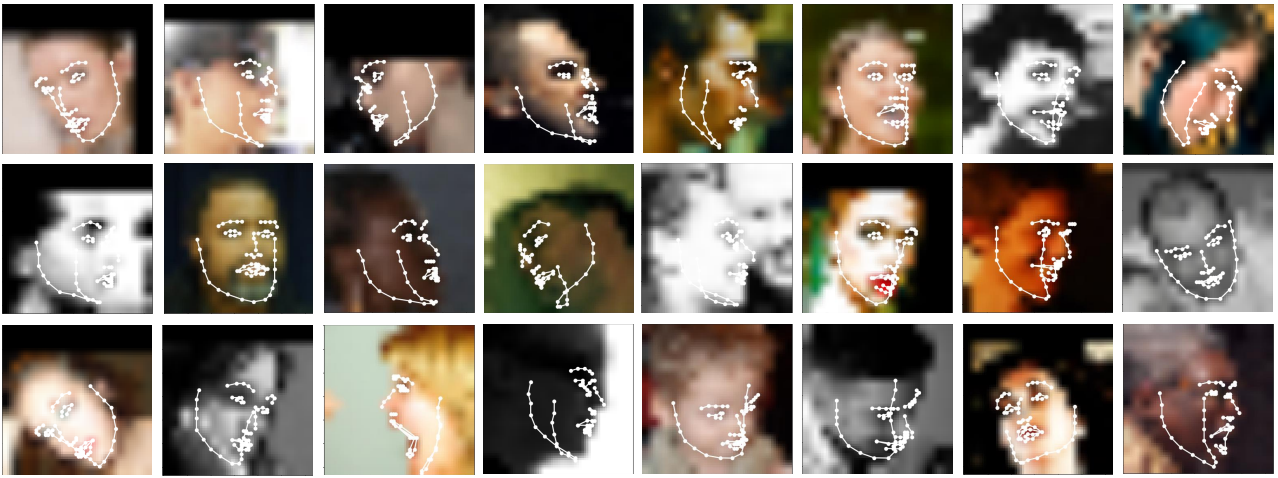

Figure 10: Fitting examples produced by Super-FAN on a few images from LS3D-W. The predictions are plotted over the original low-resolution images. Notice that our method works well for faces found in challenging conditions such as large poses or extreme illumination conditions despite the poor image quality.

14